%% file: main.tex
\documentclass[11pt]{article}

\usepackage[final]{acl}

\usepackage{times}
\usepackage{latexsym}

\usepackage[T1]{fontenc}

\usepackage[utf8]{inputenc}

\usepackage{microtype}

\usepackage{inconsolata}

\usepackage{graphicx}

\usepackage{amsmath}
\usepackage{amssymb}
\usepackage{mathtools}
\usepackage{amsthm}
\usepackage{microtype}
\usepackage{graphicx}
\usepackage{subcaption}
\usepackage{booktabs} %
\usepackage{pifont} %
\usepackage[table]{xcolor} %
\usepackage{colortbl}      %
\usepackage{subcaption}
\usepackage[capitalize,noabbrev]{cleveref}
\usepackage{multirow}

\definecolor{swablue}{RGB}{159, 195, 224}

\definecolor{swagold}{RGB}{255, 240, 180}

\definecolor{swagrey}{RGB}{239, 239, 239}

\title{Architecture-Conditioned Reinforcement Learning Makes Sliding-Window Attention Competitive in Math Reasoning}

\author{
  Kai Liu\textsuperscript{1,2}\thanks{Corresponding authors: liukai1998@tongji.edu.cn, chenkai@pjlab.org.cn.}
  \quad Peijie Dong\textsuperscript{3}
  \quad Xinchen Xie\textsuperscript{2}
  \quad Jianfei Gao\textsuperscript{2} \\
  {\bf Qipeng Guo}\textsuperscript{2} 
  \quad {\bf Xiaowen Chu}\textsuperscript{3}
  \quad {\bf Shaoting Zhang}\textsuperscript{2}
  \quad {\bf Kai Chen}\textsuperscript{2}\footnotemark[1] \\
  \textsuperscript{1}Shanghai Research Institute for Intelligent Autonomous Systems, Tongji University \\
  \textsuperscript{2}Shanghai AI Laboratory \\
  \textsuperscript{3}Hong Kong University of Science and Technology (Guangzhou) \\
}

\begin{document}
\maketitle

\input{content.tex}

\bibliography{content}

\appendix

\input{appendix.tex}

\end{document}

%% file: content.tex
\begin{abstract}
    The rapid progress of reasoning and agentic large language models (LLMs) has increased the demand for long-context inference, but self-attention (SA) scales quadratically with context length.
    To address this, we study SWARR (Sliding-Window Attention with Reinforced Adaptation for Math Reasoning), a practical recipe for adapting SWA models to mathematical reasoning. SWARR has two stages: (1) efficient conversion from a pretrained SA model to SWA with supervised fine-tuning (SFT), which avoids pretraining a new base model, and (2) policy adaptation with reinforcement learning (RL).
    We find that SWA still underperforms SA after SFT, and we hypothesize that this gap is caused in part by a data-architecture mismatch: most SFT data are prepared for SA models and may contain long-range dependencies that are difficult for SWA to model. Because on-policy RL optimizes self-generated trajectories under the SWA constraint, it can adapt trajectories to better match SWA.
    Experiments on mathematical reasoning benchmarks show that this recipe substantially narrows the gap between SWA and SA, recovering much of the accuracy lost during SWA conversion while preserving the efficiency benefits of linear-complexity attention. Our central contribution is the empirical finding that RL changes the conclusion one would draw from conversion and SFT alone about SWA's viability for math reasoning.
    Code, training and evaluation configurations, and the released checkpoint are available at \url{https://github.com/LKJacky/architecture-conditioned-rl}.
\end{abstract}

\section{Introduction}
\begin{figure*}
    \centering
    \includegraphics[width=0.95\linewidth]{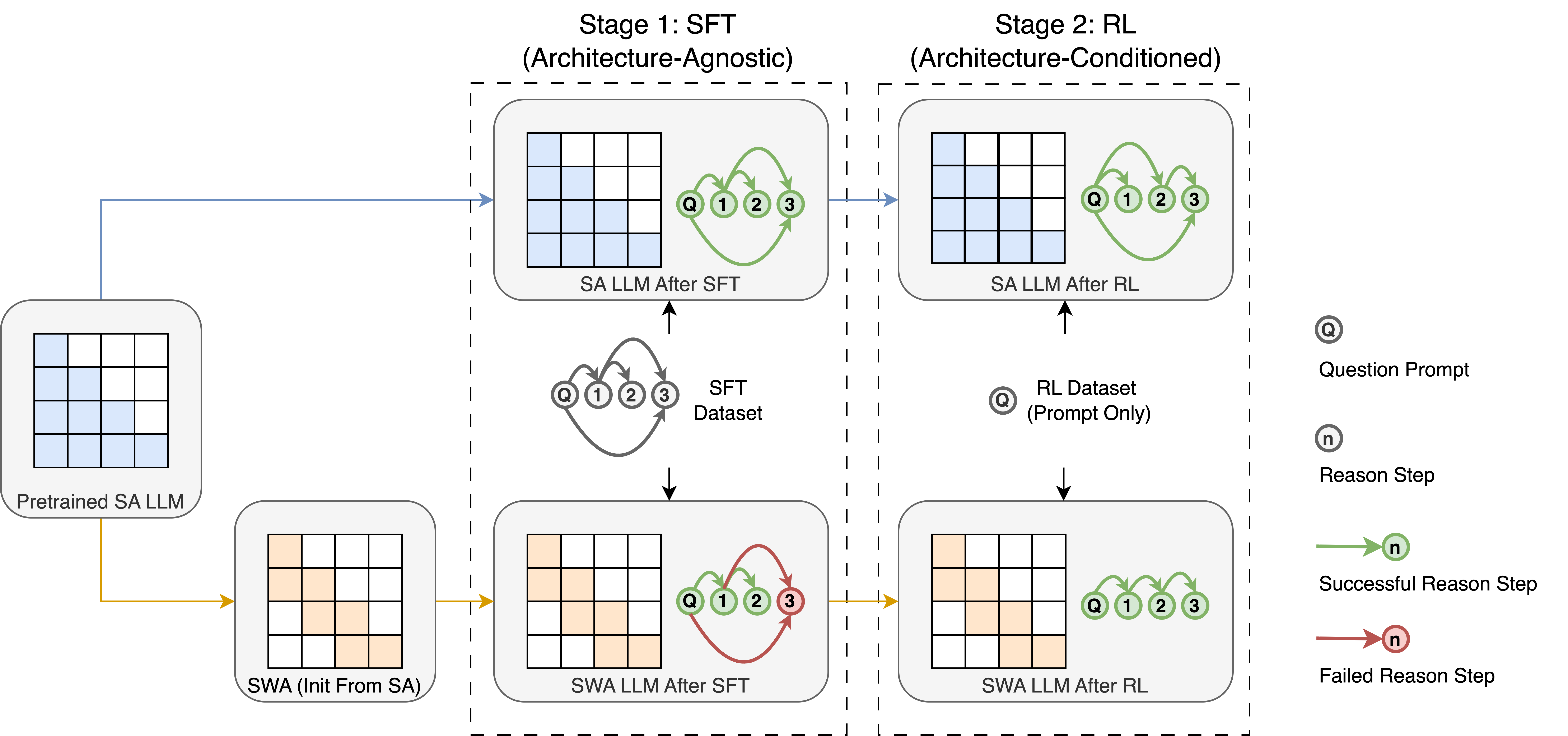}
    \caption{Overview of the \emph{SWARR} pipeline. \textbf{Stage 1:} Efficient conversion with architecture-agnostic SFT, avoiding costly pretraining. Because current SFT data are mainly collected for SA models, they may contain long-range dependencies that are difficult for SWA to model. \textbf{Stage 2:} RL policy adaptation under the SWA constraint, which may mitigate the resulting data-architecture mismatch.
    }
    \label{fig:pipeline}
\end{figure*}

Large language models (LLMs) have made substantial progress in reasoning and agentic capabilities.
ChatGPT-o1~\citep{openai2026openaio1card} exceeds human PhD-level accuracy on GPQA, a benchmark of physics, biology, and chemistry problems, by using long chain-of-thought (CoT) reasoning~\citep{wei2022cot}.
Recent code agents~\citep{team2025kimi,Chen2025MiniMaxM1ST} can solve complex software engineering problems~\citep{swebench} through multi-turn interactions.
These capabilities increasingly rely on long contexts, creating a fundamental efficiency challenge for transformer-based LLMs because self-attention (SA) has quadratic computational complexity and linear space requirements~\citep{vaswani2017attention}.

Linear-complexity attention mechanisms offer a promising direction for scaling long-context inference.
Representative approaches include recurrent neural networks (RNNs)~\citep{mamba,dao2024mamba2,yang2024gateddeltanet} and sliding-window attention (SWA)~\citep{beltagy2020longformer}.
Recent architectural innovations~\citep{mamba,peng2025rwkv,yang2024gateddeltanet} have substantially improved RNNs, making them competitive with SA on short-context tasks such as general language understanding~\citep{eval_of_mamba}.
However, because these models compress or restrict accessible memory, they still struggle to match SA on long-context tasks.
Other work, such as SmoothReading~\citep{smooth_reading}, improves linear-complexity models on long-context understanding (long input, short output) through agentic inference.
In contrast, short-input, long-generation tasks, such as mathematical reasoning, remain challenging because the model must generate extended trajectories while retaining the information needed for later reasoning steps.
This paper therefore studies whether larger-window SWA can become competitive with SA on mathematical reasoning, a representative short-input, long-generation reasoning task.

The first challenge is obtaining a strong linear-complexity base model. High-performance linear-complexity checkpoints are rare, and pretraining them from scratch is costly. We therefore revisit SWA as a practical architecture for math reasoning. SWA shares the same weight structure as SA, remains architecturally close to it, and can use efficient kernels such as FlashAttention~\citep{dao2022flashattention}. This makes it natural to convert a pretrained SA model to SWA by replacing global attention with a local sliding window and then applying supervised fine-tuning (SFT)~\citep{sft_survey}. In our experiments, this conversion stage is efficient and provides a reasonable starting point.

The second challenge is that SWA still trails SA after conversion and SFT, especially with smaller windows. A natural conclusion would be that SWA is unsuitable for long-generation reasoning. We propose a complementary explanation: a data-architecture mismatch. Standard SFT is architecture-agnostic, while most reasoning data are prepared for SA models and may contain long-range dependencies that are difficult for SWA to reproduce. This suggests that adapting the training trajectories to better match the SWA constraint could improve performance.
Reinforcement learning (RL)~\citep{schulman2017ppo} is well suited to this adaptation because on-policy optimization uses trajectories generated by the current model under its own architectural constraint. In this descriptive sense, RL is architecture-conditioned.
This motivates SWARR (Sliding-Window Attention with Reinforced Adaptation for Math Reasoning), a practical two-stage recipe including SA-to-SWA conversion with SFT and subsequent RL adaptation. On Qwen2.5-Math-1.5B-Base~\citep{qwen2_5_math}, SFT leaves SWA8k-SFT, SWA4k-SFT, and SWA2k-SFT 6.1, 9.0, and 17.8 points below SA-SFT, respectively. After RL, SWA8k comes within 0.4 points of SA under equal-step training, SWA4k becomes competitive under equal-time training, and SWA2k still lags behind, though the gap narrows to 2.5 points. The pipeline is shown in \Cref{fig:pipeline}.

To better understand this improvement, we design locality metrics for generated trajectories and conduct a controlled cross-SFT experiment using both SA- and SWA-generated data. The results show that RL is associated with more local generation that better fits the SWA architecture. In addition, SA-generated SFT data hurt SWA more than architecture-matched SWA-generated data, substantially narrowing the gap and supporting the data-architecture mismatch hypothesis. Our central contribution is therefore empirical: RL changes the conclusion one would draw from conversion and SFT alone about the viability of larger-window SWA for math reasoning.
In summary, our contributions are as follows:
\begin{itemize}
    \item We revisit SWA as an efficient linear-complexity architecture for math reasoning and show that converting a pretrained SA model with SFT provides a reasonable starting point.
    \item We present SWARR as a practical recipe for adapting SWA models to mathematical reasoning, and show that larger-window SWA substantially narrows the gap to SA.
    \item We provide analyses showing that RL is associated with more local generation and that architecture-matched SFT data improve SWA, consistent with the data-architecture mismatch hypothesis.
\end{itemize}

\section{Related Work}

\subsection{Linear-Complexity LLMs}

Linear-complexity LLMs aim to reduce the quadratic cost of self-attention in transformer-based models.
Representative approaches replace global self-attention with more efficient mechanisms, including recurrent neural networks (RNNs)~\citep{katharopoulos2020transformers,choromanski2020rethinking_performer,qin2024lightning,chou2024metala,yang2024parallelizing,yang2023gated,scaling_state_size} and sliding-window attention (SWA)~\citep{beltagy2020longformer,Jiang2023Mistral7}. Although they improve efficiency, they typically restrict accessible context and therefore still struggle on tasks that require long-range dependency modeling.

Prior work mainly addresses this limitation in two ways. \textbf{Architecture-level optimization} improves memory capacity through new architectures such as S4~\citep{s4}, Mamba~\citep{mamba}, RWKV~\citep{peng2025rwkv,peng2024eagle_rwkvv5v6}, RetNet~\citep{sun2023retentive}, TTT~\citep{sun2024ttt}, HGRN~\citep{qin2022hgrn,Qin2024HGRN2GL}, Longhorn~\citep{liu2024longhorn}, and GatedDeltaNet~\citep{yang2024gateddeltanet}.
\textbf{Data-level optimization} instead reduces the amount of long-range dependency required by the input or generation. For example, SmoothReading~\citep{smooth_reading} uses agentic multi-round inference to avoid feeding the entire context at once, improving linear-complexity LLMs on long-context understanding (long input, short output). Data-level and architecture-level optimization are complementary.
Our work extends this data-level perspective to short-input, long-generation reasoning by using RL to adapt generated trajectories to better match SWA.

\textbf{Other efficient architectures.}
Beyond linear-complexity attention, hybrid attention mechanisms~\citep{Lieber2024JambaAH,Chen2025MiniMaxM1ST,openai2025gptoss120bgptoss20bmodel,team2025hunyuan,blakeman2025nemotron,Basant2025NVIDIANemotronNano2,Gu2025JetNemotronEL,wang2025systematic,Wang2025M1TS} combine multiple attention patterns to balance efficiency and performance, while dynamic sparse attention~\citep{yuan2025nsa,lu2025moba,xiao2024infllm} restricts the attended keys through sparse patterns.
These approaches are related in goal but retain higher complexity than the linear-complexity models studied here: hybrid attention still includes quadratic components, and dynamic sparse attention has sub-quadratic computation with linear space requirements. We therefore do not treat them as primary baselines.

\subsection{Reinforcement Learning for LLMs}

Reinforcement learning with verifiable rewards (RLVR) has become an effective approach for improving LLM reasoning.
DeepSeek-R1~\citep{deepseekr1}, for example, applies GRPO~\citep{Shao2024DeepSeekMathPT} to a SA model and reports strong gains on reasoning benchmarks.
Following this line, DAPO~\citep{yu2025dapo}, GRESO~\citep{zheng2025greso}, SimpleRL-Zoo~\citep{Zeng2025SimpleRLZooIA}, and Seed-GRPO~\citep{chen2025seed} further improve RL training recipes for reasoning models.
Most existing RLVR studies mainly focus on improving the RL objective or training recipe, assuming a fixed underlying architecture, which is typically self-attention.

The interaction between RLVR and attention architecture remains underexplored. We introduce SWA as a new optional architecture for RLVR and analyze the architecture-conditioned property of RL: because RL optimizes trajectories generated by the model itself, the optimization process is constrained by the model architecture and can shift the training distribution toward patterns that better fit the underlying attention mechanism.

\subsection{Math Reasoning}

Mathematical reasoning benchmarks such as AIME24 and AIME25~\citep{aime24,aime25} require models to generate extended reasoning trajectories before producing final answers, making them representative short-input, long-generation reasoning tasks.
Because these trajectories can be long, the strong memory capacity of SA is often viewed as a key factor behind its success.

Our results provide a complementary perspective: with RL adaptation, larger-window SWA can achieve performance comparable to SA on mathematical reasoning tasks, indicating that effective reasoning does not always require unrestricted long-range attention. Instead, the generated trajectories can become more local in ways that better fit the SWA architecture.
This finding is consistent with the ``locality of thought'' phenomenon reported by \citet{locality_of_reason}, which suggests that CoT reasoning often benefits from locality in the high-level structure of the reasoning graph. Our work complements this perspective by showing that locality can also be beneficial at the low-level token sequence: effective reasoning can still emerge when the model relies on more local token dependencies.

\section{Method}

This section presents SWARR, a two-stage empirical recipe that converts a pretrained self-attention transformer into an efficient sliding-window model and then adapts its reasoning trajectories to the new attention constraint.
The overall pipeline is illustrated in \Cref{fig:pipeline}.

\subsection{Stage 1: Efficient Conversion with SFT}

The first stage initializes SWA from a pretrained SA model.
Because SWA keeps the same parameter structure as SA, we directly reuse the pretrained parameters, such as the query, key, and value (QKV) projections.
The main architectural change is replacing global attention with localized sliding-window attention, which restricts each token to attend only to a fixed-size neighborhood.

This change alters the model's inductive bias and therefore requires adaptation.
We perform SFT on mathematical reasoning data so that the converted model can recover from the attention replacement and learn to reason under its local receptive field.
For a question \(q\) and target reasoning trajectory \(o=(o_1,\dots,o_L)\), the objective is the standard cross-entropy loss:
\begin{equation}
    \small
    \text{CE}(\theta) = \mathbb{E}_{(q,o) \sim \mathcal{D}_{\text{SFT}}} \left[ - \sum_{t=1}^L \log \pi_{\text{SWA}}(o_t|q, o_{<t}) \right]
\end{equation}
For fair comparison, we apply the same SFT data and training settings to the SA baseline.

\textbf{Architecture-agnostic nature of SFT.}
SFT uses fixed target trajectories, so the data remain unchanged regardless of the underlying attention mechanism.
Existing SFT datasets are typically prepared for SA models and may contain dependencies that extend beyond the SWA window.
As a result, SWA is asked to imitate trajectories that may require information unavailable under its local attention pattern, leading to suboptimal performance after SFT.
This data-architecture mismatch motivates the second stage, where RL may adapt the generation process to better fit SWA.

\subsection{Stage 2: Policy Adaptation with RL}

Starting from the SFT-converted SWA model, we apply RL to improve mathematical reasoning and shift the model toward trajectories that are easier to generate with local attention.
We use the CISPO~\citep{Chen2025MiniMaxM1ST} loss:
\begin{equation}
    \small
    \begin{aligned}
         & \mathcal{J}_{\mathrm{CISPO}}(\theta)
        =
        \mathbb{E}_{(q),\,\{o_i\}_{i=1}^{G}}
        \Bigg[
            \frac{1}{\sum_{i=1}^{G}|o_i|}
            \sum_{i=1}^{G}
            \sum_{t=1}^{|o_i|}
        \\
         & \qquad
            \mathrm{sg}\!\left(\hat{r}_{i,t}(\theta)\right)
            \hat{A}_{i,t}
            \log \pi_{\theta}
            \!\left(o_{i,t}\mid q,o_{i,<t}\right)
        \Bigg]                                                                                               \\
         & \quad where\ (q,a)\!\sim\!\mathcal{D},\ \{o_i\}_{i=1}^{G}\!\sim\!\pi_{\mathrm{old}}(\cdot\mid q), \\
         & \qquad \hat{r}_{i,t}(\theta)
        =
        \mathrm{clip}\!\left(
        r_{i,t}(\theta),
        1-\epsilon^{IS}_{\mathrm{low}},
        1+\epsilon^{IS}_{\mathrm{high}}
        \right)
    \end{aligned}
\end{equation}
Here, \(q\) denotes a question, \(\{o_i\}_{i=1}^{G}\) are sampled trajectories from the old policy, \(\hat{r}_{i,t}(\theta)\) is the clipped importance-sampling ratio, and \(\hat{A}_{i,t}\) is the group-normalized advantage.

\textbf{Architecture-conditioned characteristic of RL.}
Unlike SFT, RL trains on outputs sampled from the current model, \(o\sim\pi_\mathrm{old}(\cdot|q)\), rather than fixed trajectories \(o\sim\mathcal{D}_{\mathrm{SFT}}\).
Therefore, the training distribution changes with the model architecture and policy.
For SWA, this property allows RL to reinforce successful trajectories sampled under the sliding-window constraint, which may favor more local dependencies that better fit the architecture.
This adaptation may help mitigate the performance drop caused by the mismatch between SA-oriented SFT data and the SWA architecture.

\newcommand{\timeof}[1]{\(\text{T}_\text{#1}\)}
\newcommand{\metric}[2]{\ensuremath{#1_{\pm #2}}}
\begin{table*}[t]
    \centering
    \small
    \caption{Main results and compute cost. The table is organized into three blocks: the first compares SA and SWA models after SFT, the second compares models trained for the same number of RL steps, and the third compares models trained for similar RL training time. \timeof{Train} reports SFT GPU hours in the SFT block and additional RL GPU hours in the RL blocks; the total training cost of an RL model is the sum of its SFT cost and the reported RL cost. \timeof{Eval} denotes the evaluation GPU hours for one full benchmark run. Accuracy is reported as \(\mathrm{score}_{\pm h}\), where \(h\) is the half-width of the 95\% bootstrap confidence interval.}\label{tab:main_results}
    \begin{tabular}{l|cc|ccccc|c|c}
        \toprule
        Model         & \timeof{Train} & \timeof{Eval} & Math500            & AMC                & Olymp              & Aime24             & Aime25             & Avg                & Gap      \\
        \midrule
        \multicolumn{10}{c}{\cellcolor{swagrey} \texttt{SFT Models}}                                                                                                                            \\
        \midrule
        SA-SFT        & 615            & 10.44         & \metric{75.0}{1.4} & \metric{50.3}{2.1} & \metric{45.5}{1.2} & \metric{40.1}{2.2} & \metric{32.3}{2.0} & \metric{48.6}{0.8} & baseline \\
        \midrule
        SWA8k-SFT     & 560            & 5.54          & \metric{68.9}{1.6} & \metric{41.3}{2.2} & \metric{41.8}{1.2} & \metric{34.2}{2.3} & \metric{26.4}{2.0} & \metric{42.5}{0.8} & -6.1     \\
        SWA4k-SFT     & 528            & 4.70          & \metric{70.3}{1.6} & \metric{42.0}{2.2} & \metric{41.5}{1.2} & \metric{22.4}{2.0} & \metric{21.8}{1.9} & \metric{39.6}{0.8} & -9.0     \\
        SWA2k-SFT     & 507            & 4.20          & \metric{63.1}{1.5} & \metric{32.7}{2.0} & \metric{33.2}{1.1} & \metric{13.8}{1.8} & \metric{11.1}{1.6} & \metric{30.8}{0.7} & -17.8    \\
        \midrule
        \multicolumn{10}{c}{\cellcolor{swagrey} \texttt{RL Models with 900 Steps}}                                                                                                              \\
        \midrule
        SA-RL-900     & 498            & 2.81          & \metric{89.1}{0.8} & \metric{72.4}{2.1} & \metric{61.4}{1.0} & \metric{56.6}{2.2} & \metric{50.2}{2.1} & \metric{65.9}{0.8} & baseline \\
        \midrule
        SWA8k-RL-900  & 337            & 2.11          & \metric{89.4}{0.9} & \metric{71.1}{2.2} & \metric{61.3}{1.0} & \metric{56.6}{2.2} & \metric{49.3}{2.2} & \metric{65.5}{0.8} & -0.4     \\
        SWA4k-RL-900  & 285            & 1.84          & \metric{88.6}{0.9} & \metric{70.2}{2.1} & \metric{61.6}{1.0} & \metric{53.0}{2.2} & \metric{44.2}{2.1} & \metric{63.5}{0.8} & -2.4     \\
        SWA2k-RL-900  & 225            & 1.46          & \metric{87.1}{1.0} & \metric{66.1}{1.9} & \metric{59.0}{1.0} & \metric{46.8}{2.2} & \metric{39.2}{2.0} & \metric{59.6}{0.8} & -6.3     \\
        \midrule
        \multicolumn{10}{c}{\cellcolor{swagrey} \texttt{RL Models with about 500 GPU Hours RL Training}}                                                                                        \\
        \midrule
        SWA8k-RL-1200 & 466            & 2.23          & \metric{89.3}{0.8} & \metric{71.4}{2.0} & \metric{63.1}{1.1} & \metric{60.3}{2.2} & \metric{49.1}{2.1} & \metric{66.6}{0.8} & +0.7     \\
        SWA4k-RL-1400 & 474            & 1.87          & \metric{89.4}{0.8} & \metric{74.5}{2.0} & \metric{62.1}{1.0} & \metric{57.5}{2.1} & \metric{46.4}{2.0} & \metric{66.0}{0.8} & +0.1     \\
        SWA2k-RL-1700 & 470            & 1.33          & \metric{88.5}{0.9} & \metric{71.7}{2.1} & \metric{61.0}{1.0} & \metric{52.4}{2.1} & \metric{43.5}{2.0} & \metric{63.4}{0.8} & -2.5     \\
        \bottomrule
    \end{tabular}
\end{table*}

\subsection{Measuring Locality of Generation}

To verify the hypothesis that RL adaptation leads to more local generation, we design metrics to measure the locality of generated trajectories.
To measure how much predictive information lies within a fixed window, we compute a probability-based information gap.
For a token sequence \(x_{1:T}\) and window size \(w\), we define
\begin{equation}
    \small
    g_t(w)
    =
    p(x_t \mid x_{<t})
    -
    p(x_t \mid x_{t-w:t-1})
    \qquad t>w .
\end{equation}
Here, probability \(p\) is measured by a reference model. The first term uses the full prefix, while the second uses only the recent \(w\) tokens.
\(g_t(w)\) measures the information gap between full-context and local-window for next-token prediction.
Smaller \(g_t(w)\) indicates that the local window captures more predictive information.
We then define the sequence-level locality as:
\begin{equation}
    \small
    L(w)
    =
    1-\frac{1}{T-w}
    \sum_{t=w+1}^{T} g_t(w).
\end{equation}
\(L(w)\) measures how much predictive information is captured by a local window relative to the full context. Higher values indicate greater reliance on local context and hence better alignment with SWA.
For efficiency, when using stride \(s>1\), we group \(s\) consecutive target tokens into one segment; this slightly expands the available context within each group and therefore yields a lower-bound approximation of the exact locality gap.

\section{Experiments}

\subsection{Experimental Setup}\label{sec:exp_setup}

\textbf{Model.} We use Qwen2.5-Math-1.5B-Base~\citep{qwen2_5_math} as the SA base model, denoted SA-Base. We convert the same checkpoint into SWA2k/4k/8k-Base. Unless noted otherwise, all models are 1.5B. For fair comparison, SA and SWA share the same SFT and RL recipes; the only intended difference is the attention architecture.

\textbf{SFT settings.}
We use a 42B-token math-heavy dataset and train for one epoch with batch size 1M tokens (42,426 steps), learning rate \(8\times10^{-5}\), weight decay 0.1, cosine decay, 3\% warmup, and 128k maximum sequence length.

\textbf{RL settings.}
We use AceReason-Math~\citep{liu2025acereason} (49k samples), a batch size of 128 prompts with 8 rollouts each, a learning rate of \(10^{-6}\), weight decay 0.1, and 4 optimization updates per rollout batch, with asymmetric clipping set to \texttt{h-clip}=0.28 and \texttt{l-clip}=0.2.

\textbf{Evaluation.}
We evaluate on AMC~\citep{amc}, MATH500~\citep{hendrycks2021measuring}, OlympiadBench~\citep{he2024olympiadbench}, AIME24~\citep{aime24}, and AIME25~\citep{aime25}. These benchmarks are repeated 8, 2, 2, 32, and 32 times, yielding 664, 1000, 1348, 960, and 960 samples, respectively.

The maximum generation length is 64k for both RL training and evaluation. More details on the experimental setup are provided in \Cref{sec:appendix_experiment_details}.

\begin{figure*}
    \centering
    \includegraphics[width=\textwidth]{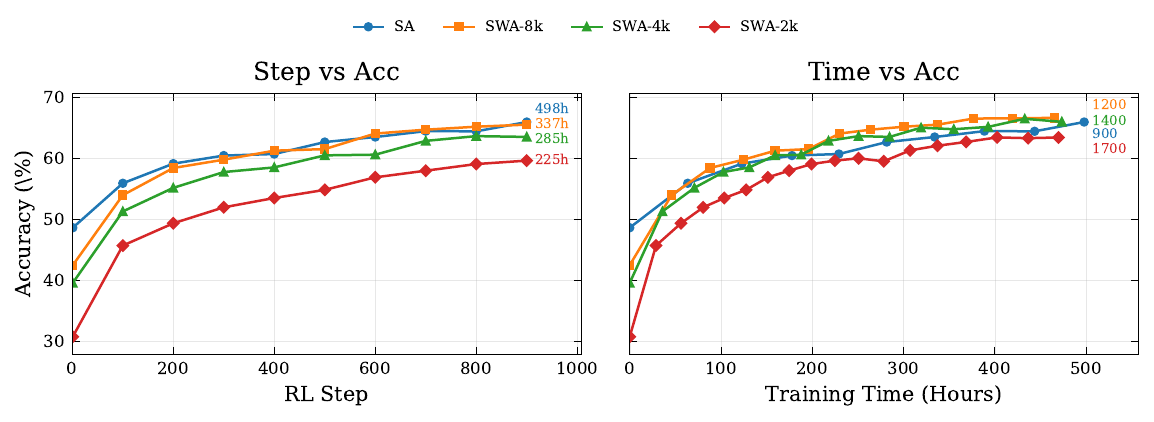}
    \caption{
        RL training curves for SA and SWA models. Panel (left) reports accuracy as a function of RL steps, while panel (right) reports accuracy as a function of actual training time, highlighting the efficiency advantage of SWA.
    }
    \label{fig:rl_curve}
\end{figure*}

\newcommand{\dataof}[1]{\(\text{D}_{\text{#1}}\)}

\subsection{Results After SFT}

We first train SA-Base, SWA2k-Base, SWA4k-Base, and SWA8k-Base with the same SFT recipe.
The results are shown in the first block of Table~\ref{tab:main_results}.

\textbf{Finding 1: SWA achieves reasonable mathematical reasoning performance after SFT, but still underperforms SA, and the gap increases as the window size decreases.}
SWA8k-SFT still reaches 42.5\% average accuracy after conversion and SFT, providing a strong starting point. However, SWA remains below SA, and the gap widens as the window shrinks: SWA8k-SFT, SWA4k-SFT, and SWA2k-SFT trail SA by 6.1, 9.0, and 17.8 points, respectively. This monotonic trend suggests that smaller windows make SA-oriented SFT trajectories harder to imitate.

\textbf{Finding 2: SWA models are more efficient than SA.}
SWA reduces both training and evaluation cost. For example, SWA8k-SFT cuts evaluation time from 10.44 to 5.54 GPU hours relative to SA-SFT, and smaller windows reduce cost further.

\subsection{Results After RL}

The RL results are shown in the second and third blocks of Table~\ref{tab:main_results} with training curves in \Cref{fig:rl_curve}.
We report two complementary comparisons: the second block controls for the number of RL updates by training all models for the same 900 RL steps, while the third block controls for RL training time by training SWA models for about 500 GPU hours, which is a practical situation in which the training budget is limited.

\textbf{Finding 3: SWA models can train more steps within the same training-time budget because of cheaper rollouts.}
Because SWA rollouts are cheaper, the similar-time setting allows more RL steps. Under this comparison, SWA8k-RL-1200 and SWA4k-RL-1400 are trained for about 500 GPU hours, which is 33\% and 56\% more steps than SA-RL-900, respectively. This advantage gives SWA more RL adaptation within the same training-time budget.

\textbf{Finding 4: RL substantially improves SWA, making larger-window models competitive with SA while smaller-window models remain more constrained.}
Under the same-step comparison, SWA8k-RL-900 achieves performance comparable to SA-RL-900, with an average accuracy gap of only -0.4\%. Under the similar-time comparison, SWA4k-RL-1400 also reaches comparable performance to SA-RL-900, with a small positive gap of +0.1\%. We do not interpret small positive gaps such as +0.7 and +0.1 as evidence of clear superiority, because they are within repeated-evaluation uncertainty.
In contrast, the gap of SWA2k decreases to -6.3\% under the same-step comparison and -2.5\% under the similar-time comparison, but it still underperforms SA, suggesting that very small windows impose a memory constraint that is harder to overcome with RL.

\section{Analysis}

To understand why RL narrows the SA-SWA gap, we analyze SWA generation, test the data-architecture mismatch hypothesis through a cross-SFT experiment, ablate SWA conversion, and study the inference efficiency of SWA.

\subsection{Characteristics of SWA Generation}

\subsubsection{Generation Length}

\textbf{Finding 5: SWA generates shorter trajectories than SA, but correct SWA trajectories can still extend far beyond the attention window.}
We measure the mean output length of trajectories and the percentage of correct samples whose output length exceeds the SWA window size on AIME24, as shown in \Cref{tab:length}.
Mean output length decreases for all models during RL, but more sharply for SWA. At step 900, the mean lengths are 14,669, 12,601, 12,628, and 10,731 for SA-RL-900, SWA8k-RL-900, SWA4k-RL-900, and SWA2k-RL-900, respectively. Smaller windows therefore appear to favor shorter trajectories.
At the same time, many correct SWA samples still exceed the attention window: 42.91\%, 78.98\%, and 99.78\% for SWA8k/4k/2k-RL-900, respectively. This suggests that SWA can still sustain long reasoning chains.

\begin{figure*}[!htbp]
    \centering
    \includegraphics[width=0.99\textwidth]{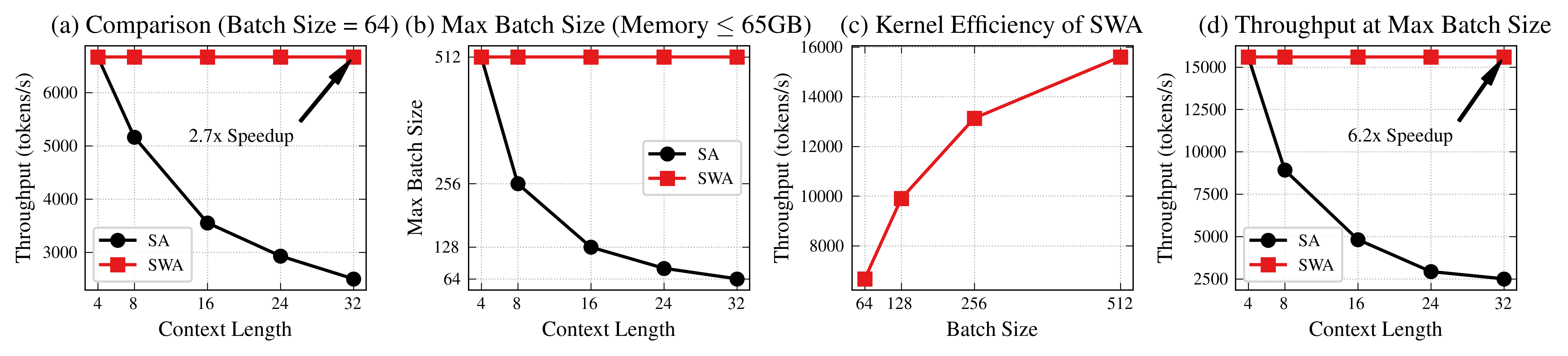}
    \caption{Efficiency comparison of SA and SWA4k. (a) Throughput as a function of context length. (b) Maximum batch size supported under a 65GB memory constraint. (c) Throughput as a function of batch size, where larger batch size leads to higher kernel efficiency. (d) Throughput with the maximum supported batch size, showing that SWA4k achieves an approximately 6.2$\times$ speedup over SA at 32k context length.}
    \label{fig:efficiency}
\end{figure*}

\newcommand{\aofb}[2]{\(\text{#1}_{\text{#2}}\)}
\begin{table}[t]
    \centering
    \small
    \caption{Generation-length analysis. \aofb{Length}{n} denotes the mean output length at the \(n\)-th RL step, and \aofb{R}{>w} denotes the percentage of correct samples whose output length exceeds the SWA window size. \aofb{R}{>w} is calculated at the 900th RL step.}
    \label{tab:length}
    \begin{tabular}{l|cc|c}
        \toprule
        Model & \aofb{Length}{0} & \aofb{Length}{900} & \aofb{R}{>w} \\
        \midrule
        SA    & 36382            & 14669              & /            \\
        SWA8k & 36532            & 12601              & 42.91\%      \\
        SWA4k & 39284            & 12628              & 78.98\%      \\
        SWA2k & 38913            & 10731              & 99.78\%      \\
        \bottomrule
    \end{tabular}
\end{table}

\begin{table}[t]
    \centering
    \small
    \caption{Probability-based locality gap \(L(w)\) on AIME24 under matched length distributions. Larger values indicate that a fixed local window captures more of the predictive information needed for next-token prediction.}
    \label{tab:content_gap}
    \begin{tabular}{l|ccc}
        \toprule
        Trajectory   & L(2048)        & L(4096)        & L(8192)        \\
        \midrule
        SA-RL-900    & 94.12          & 95.97          & 98.35          \\
        \midrule
        SWA8k-RL-900 & 95.88          & 97.55          & 99.29          \\
        SWA4k-RL-900 & 96.71          & 98.42          & 99.61          \\
        SWA2k-RL-900 & \textbf{97.65} & \textbf{99.00} & \textbf{99.90} \\
        \bottomrule
    \end{tabular}
\end{table}

\begin{table}[t]
    \centering
    \small
    \caption{Performance comparison under different training-data settings.
        \dataof{SA}, \dataof{SWA8k}, and \dataof{SWA4k} datasets are generated from SA-RL-1300, SWA8k-RL-1500, and SWA4k-RL-2200 respectively, with only correct trajectories and matched length distributions via resampling.
        All source models are best-performing checkpoints for their respective architectures within 2500 RL steps, achieving similar performance with 68.1, 68.6, and 68.2 average accuracy.
    }
    \label{tab:cross_sft}
    \begin{tabular}{lcc}
        \toprule
        \multicolumn{3}{c}{\cellcolor{swagrey}\dataof{SA} vs \dataof{SWA8k}} \\
        \midrule
        Base Model         & \dataof{SA}        & \dataof{SWA8k}             \\
        \midrule
        SA-Base            & \metric{55.5}{0.7} & \metric{57.7}{0.7}         \\
        SWA8k-Base         & \metric{53.8}{0.8} & \metric{58.2}{0.7}         \\
        \midrule
        Gap (SWA8k $-$ SA) & -1.7               & +0.5                       \\
        \midrule
        \multicolumn{3}{c}{\cellcolor{swagrey}\dataof{SA} vs \dataof{SWA4k}} \\
        \midrule

        Base Model         & \dataof{SA}        & \dataof{SWA4k}             \\
        \midrule
        SA-Base            & \metric{55.5}{0.7} & \metric{57.5}{0.7}         \\
        SWA4k-Base         & \metric{49.1}{0.7} & \metric{56.8}{0.7}         \\
        \midrule
        Gap (SWA4k $-$ SA) & -6.4               & -0.7                       \\
        \bottomrule
    \end{tabular}
\end{table}

\begin{table}[t]
    \small
    \centering
    \caption{Comparison of conversion strategies for SWA2k. Math Avg denotes average accuracy across mathematical reasoning benchmarks.}\label{tab:training_time_comparison}
    \begin{tabular}{l|cc|c}
        \toprule
        Model        & Base Param & SFT & Math Avg      \\
        \midrule
        SWA2k-Random & Random     & Yes & 3.00          \\
        SWA2k-No-SFT & SA-SFT     & No  & 20.5          \\
        \midrule
        SWA2k-SFT    & SA-Base    & Yes & \textbf{30.8} \\
        \bottomrule
    \end{tabular}
\end{table}

\subsubsection{Locality of Generation}

To compare locality independent of length, we keep correct trajectories longer than 8k tokens and resample them so that all models have the same length distributions.
We use SA-SFT as the reference model to measure the locality \(L(w)\) of trajectories from all models; the results are shown in \Cref{tab:content_gap}.
\textbf{Finding 6: SWA trajectories after RL are more local than SA trajectories under matched length distributions.}
SWA trajectories are more local than SA trajectories, and smaller windows lead to higher locality. This is consistent with SWA favoring more local dependencies. The results also suggest that long-range dependencies are not strictly necessary for correct reasoning, and that SWA can learn to rely more on local context to achieve strong performance.

\subsection{Does SFT Always Cause a Drop?}

We test whether the SFT drop is caused by a mismatch between SA-oriented trajectories and the SWA architecture. To do so, we collect 3.3B-token SFT datasets from RL-trained SA, SWA8k, and SWA4k models. To reduce simple confounds, we keep only correct trajectories and resample the datasets so that they have the same length distribution with a 1k-token bin size. We then train SA-Base, SWA8k-Base, and SWA4k-Base on these datasets in a cross-setting for 13,000 steps with a batch size of 256k tokens. Results are shown in \Cref{tab:cross_sft}.
\textbf{Finding 7: SA-generated SFT data hurt SWA more, while architecture-matched SWA-generated data substantially narrow the gap.}
When trained on the SA-generated SFT dataset, SWA8k and SWA4k underperform SA by -1.7\% and -6.4\%, respectively.
In contrast, when trained on SWA-generated SFT datasets, the gaps shrink to +0.5\% for SWA8k data and -0.7\% for SWA4k data.
Overall, the results are consistent with our hypothesis that SA-generated trajectories contain dependencies that are less suitable for SWA, whereas architecture-matched SWA-generated trajectories better fit the local attention constraint.

\subsection{Effectiveness of Conversion from SA}

We further evaluate the effectiveness of converting from a pretrained SA model for SWA training.
We compare SWA2k-SFT with two baselines: (1) SWA2k-Random, which trains an SWA2k model from random initialization without loading pretrained SA weights, and (2) SWA2k-No-SFT, which converts from SA-SFT but does not perform additional SFT after conversion.
The results are shown in \Cref{tab:training_time_comparison}.
\textbf{Finding 8: Conversion from a pretrained SA model is essential for effective SWA training.}
SWA2k-Random reaches only 3.0\% average accuracy, far below SWA2k-SFT at 30.8\%. SWA2k-No-SFT achieves 20.5\%, which is also much lower than SWA2k-SFT.
These results show that reusing pretrained SA weights provides a stronger initialization than training SWA from scratch, and additional SFT after conversion further improves SWA performance.

\subsection{Efficiency Analysis}

Decoding dominates RL rollout cost for long-generation reasoning, so we compare SWA4k and SA under the same hardware, decoding stack, batching configuration, and kernel implementation. The only intended difference is the attention mechanism and cache policy. For SWA evaluation and timing, we use a true sliding KV cache. \Cref{fig:efficiency} shows two main advantages of SWA.
\textbf{Computational efficiency.}
As shown in \Cref{fig:efficiency}(a), SWA maintains high throughput as context length increases because its attention computation is local rather than global.
\textbf{Memory efficiency.}
Under a fixed 65GB memory budget, SWA4k supports a substantially larger maximum batch size than SA.
As shown in \Cref{fig:efficiency}(b), the maximum batch size of SA drops by a factor of 8, from 512 to 64, as context length grows, whereas SWA4k remains largely unaffected.
\Cref{fig:efficiency}(c) further shows that SWA4k throughput increases steadily with larger batch sizes, because of higher kernel efficiency and better GPU utilization.
\Cref{fig:efficiency}(d) compares throughput at the maximum supported batch sizes, SWA4k achieves an approximately 6.2$\times$ speedup over SA at 32k context length.

\textbf{End-to-end speedup decomposition.}
The 6.2$\times$ figure above measures the peak-throughput advantage of SWA at a fixed context length. In actual RL deployments, the total speedup is compounded by shorter generated trajectories: RL produces more concise outputs for SWA, so decoding fewer tokens amplifies the efficiency benefit. To separate the two factors, we measure end-to-end inference of the RL-trained models on a single GPU (Table~\ref{tab:efficiency_decomposition}). SA-RL-900 generates a mean output length of 10{,}022 tokens across the five benchmarks, whereas SWA4k-RL-900 generates 8{,}450 tokens. Holding the mean length fixed, changing only the architecture from SA to SWA4k yields a 2.1$\times$ speedup; further reducing the mean length to 8{,}450 raises the speedup to 2.5$\times$. Under this decomposition, the architecture change contributes 87.6\% of the total time saving, while length reduction contributes 12.4\%, confirming that the dominant efficiency gain comes from SWA's lower per-token attention cost rather than from shorter generations.

\begin{table*}[t]
    \centering
    \small
    \begin{tabular}{lcccc}
        \toprule
        Model & Mean Length            & Infer. GPU Hours & Speedup     & Contribution           \\
        \midrule
        SA    & 10{,}022 (SA-RL-900)   & 1.46             & 1$\times$   & ---                    \\
        SWA4k & 10{,}022 (SA-RL-900)   & 0.69             & 2.1$\times$ & 87.6\% (change arch.)  \\
        SWA4k & 8{,}450 (SWA4k-RL-900) & 0.58             & 2.5$\times$ & 12.4\% (reduce length) \\
        \bottomrule
    \end{tabular}
    \caption{End-to-end inference speedup decomposition of SWA4k over SA. Mean lengths are averaged over the five benchmarks for SA-RL-900 and SWA4k-RL-900, and inference is measured on a single GPU; the contribution column reports each factor's share of the total time saving.}
    \label{tab:efficiency_decomposition}
\end{table*}

\subsubsection{More Experiments}

Additional experiments in \Cref{sec:appendix_more_experiments} further support the main conclusions.
\begin{itemize}
    \item  \Cref{sec:appendix_text_locality} provides a text-level locality analysis showing that SWA trajectories exhibit less long-range reuse than SA trajectories under matched length distributions, consistent with the locality results in the main text.
    \item  \Cref{sec:appendix_case_study} further provides a qualitative case study comparing SA and SWA trajectories.
    \item  \Cref{sec:appendix_4b} shows that the same SFT-to-RL trend persists at 4B scale, suggesting that the benefits of architecture-conditioned RL extend across model sizes.
    \item  \Cref{sec:appendix_other_domains} evaluates SWA on other domains beyond mathematical reasoning, discussing the potential and challenges of SWA for broader tasks.
    \item  \Cref{sec:appendix_ceiling_effect} shows that the performance gap does not shrink when the SA SFT model performs worse than the SWA SFT model, ruling out the possibility that the observed gap is due to a ceiling effect.
    \item  \Cref{sec:appendix_plain_grpo} shows that the gap-shrinking trend also appears under plain GRPO without the stabilization stack, indicating that the trend is a general property of architecture-conditioned RL rather than an artifact of the stabilization techniques.
    \item  \Cref{sec:appendix_on_off_policy} compares on-policy and off-policy RL, showing that on-policy RL substantially outperforms off-policy RL for SWA and supporting the view that architecture-conditioned trajectories are important for SWA.
    \item  \Cref{sec:appendix_open_sft} further shows that this trend generalizes across datasets: on the public AceReason math SFT data~\citep{liu2025acereason}, SWA2k trails SA by 4.2 points after SFT, and after 500 RL steps it reaches 46.6\%, slightly above SA (46.2\%).
\end{itemize}

\section{Conclusion}

We present SWARR as a practical empirical recipe for adapting SWA to mathematical reasoning through SA-to-SWA conversion followed by RL adaptation. SFT alone leaves a substantial gap between SWA and SA, but RL makes larger-window SWA competitive while preserving its efficiency advantages. We view the gap-shrinking trend itself as a robust empirical finding. Our analyses are consistent with a data-architecture matching explanation: RL produces shorter, more local trajectories, and controlled cross-SFT experiments show that architecture-matched SWA-generated data substantially narrow the gap. Overall, our main contribution is the empirical finding that RL changes the conclusion one would draw from conversion and SFT alone about the viability of larger-window SWA for long-generation math reasoning.

\section*{Limitations}

Our study focuses on mathematical reasoning, which provides a clean setting for short-input, long-generation behavior. This task scope is important because different tasks may require different optimization strategies, data mixtures, or inference procedures. As a result, our results provide evidence for mathematical reasoning, but should not be viewed as a universal recipe for all tasks.
As an academic study, we evaluate 1.5B and 4B models.
We also note that validation at larger scale and on additional domains is ongoing; preliminary results along these lines are encouraging, and we leave their systematic study to future work.
Finally, we restrict our study to pure sliding-window attention without global tokens. Adding a small number of global (attention-sink style) tokens would provide a lightweight global view of the context, partially relaxing the locality constraint and potentially narrowing the remaining gap while supporting even smaller window sizes; such hybrid variants would introduce additional design choices and evaluation complexity, and we leave their investigation to future work.

%% file: appendix.tex
\section{Appendix}\label{sec:appendix}

\subsection{Acknowledgments}

This research project was supported by Shanghai Artificial Intelligence Laboratory.

\subsection{More Experiments}\label{sec:appendix_more_experiments}

\subsubsection{Locality Analysis on Text-level}\label{sec:appendix_text_locality}

To examine whether RL is associated with changes in generated reasoning traces, we measure the locality of repeated content in model outputs.
For a decoded token sequence \(x_{1:T}\) and a fixed \(n\)-gram size \(n\), define each \(n\)-gram as
\begin{equation}
    \small
    g_i = (x_i, x_{i+1}, \dots, x_{i+n-1}), \qquad i=1,\dots,T-n+1.
\end{equation}
If \(g_i\) previously appeared at position \(j<i\), its recurrence gap is
\begin{equation}
    \small
    \Delta_i = i-j.
\end{equation}

\paragraph{Long@k.}
We measure the frequency with which a model reuses content after more than \(d\) tokens:
\begin{equation}
    \small
    \mathrm{Long@}d = \frac{1000}{T} \sum_i \mathbf{1}[\Delta_i > d],
\end{equation}
This metric counts repeated \(n\)-grams whose recurrence gap exceeds \(d\), normalized per 1k generated tokens.
We report thresholds such as \(d\in\{2048,4096,\dots\}\).
Lower \(\mathrm{Long@}d\) values indicate less long-range reuse and therefore better alignment with SWA's local attention pattern.

\paragraph{Gap statistics.}
We further summarize the recurrence gaps using their median and mean:
\begin{equation}
    \small
    \begin{aligned}
         & \mathrm{gap(med)} = \mathrm{median}(\{\Delta_i\}), \\
         & \mathrm{gap(mean)} = \mathrm{mean}(\{\Delta_i\}).
    \end{aligned}
\end{equation}
Smaller gap statistics likewise indicate more local reuse and better alignment with SWA.

\begin{table}[h]
    \centering
    \small
    \caption{Locality of generation for SA-RL-900 and SWA2k-RL-900 under matched length distributions. We measure long-range dependence in generated trajectories on AIME24. We keep only samples with correct final answers and output lengths longer than 8k tokens, then resample them so that SA and SWA2k have the same length distributions with a 1k-token bin size. Lower Long@\(w\) and smaller gap statistics indicate more local reuse.
    }
    \label{tab:locality}
    \begin{tabular}{l|cc}
        \toprule
        Model        & SA-RL-900 & SWA2k-RL-900 \\
        \midrule
        length(mean) & 12518     & 12555        \\
        \midrule
        Long@2048    & 41.64     & 34.16        \\
        Long@4096    & 17.81     & 13.78        \\
        \midrule
        gap(med)     & 201.42    & 183.20       \\
        gap(mean)    & 884.29    & 804.56       \\
        \bottomrule
    \end{tabular}
\end{table}

The results are shown in \Cref{tab:locality}.
SWA2k-RL-900 has lower Long@2048 and Long@4096 values than SA-RL-900, with 34.16 vs. 41.64 and 13.78 vs. 17.81, respectively.
It also has smaller recurrence gaps, with 183.20 vs. 201.42 for gap(med) and 804.56 vs. 884.29 for gap(mean).
These results indicate that SWA relies less on long-range reuse and more on local dependencies, which is consistent with its sliding-window attention pattern.

\begin{figure}
    \centering
    \includegraphics[width=\linewidth]{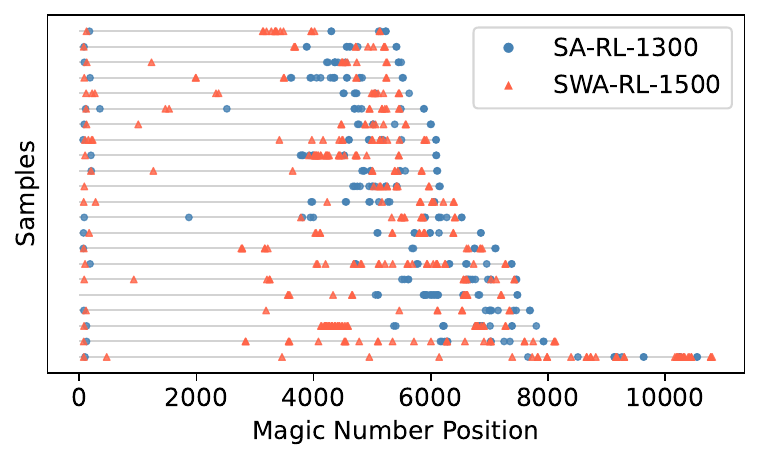}
    \caption{
        Case study comparing trajectories from SA-RL-1300 and SWA2k-RL-1500 on AIME24 with magic-number injection, matched by length distribution.
    }\label{fig:case_study}
\end{figure}

\subsubsection{Case Study of Generated Trajectories}
\label{sec:appendix_case_study}

Here, we provide a qualitative case study of generated trajectories to further illustrate the differences between SA and SWA. Tracing long generations directly is difficult, so we introduce a simple visualization based on magic-number injection.
Given an original question, we construct a modified prompt that asks the model to add a fixed magic number to the final answer: ``\emph{[original question] Then add the result and [magic number] to obtain the final answer, and place it in \textbackslash boxed\{\}.}''
Because the magic-number addition is explicitly requested at the end, we can trace where the model reintroduces this final constraint by locating the magic number in the generated trajectory.
We collect trajectories from SA-RL-1300 and SWA2k-RL-1500 on AIME24 using magic number \texttt{7898}, keeping only correct cases with matched length distributions. We first visualize the position of the magic number in each trajectory, as shown in \Cref{fig:case_study}. The magic number appears more frequently near the end of SA-RL-1300 trajectories, while it is distributed more evenly throughout SWA2k-RL-1500 trajectories.
Manual inspection suggests that SWA2k-RL-1500 more often restates the final constraint during intermediate reasoning, for example ``\emph{The problem is to find the sum of squares, and then add 7898.}'' and ``\emph{Perhaps I can calculate the sum of squares for the list, and it should be a number, and when I add 7898, it gives the answer.}'' These repetitions appear natural and may help the model keep the final requirement active during reasoning, which is qualitatively consistent with SWA's local attention pattern. Similar repetitions are also observed in SA-RL-1300, but less frequently. We emphasize that this case study is illustrative rather than causal, and we leave more systematic analysis to future work.

\subsubsection{Experiments with 4B scale}\label{sec:appendix_4b}

To further evaluate SWA at larger scales, we conduct experiments with 4B-parameter models. The base model is Qwen3-4B-Base~\citep{qwen3}, and the results are shown in \Cref{tab:4b_results}. SWA12k-4B-SFT achieves 80.2\% average accuracy, 3.9 points below SA-4B-SFT at 84.1\%. After RL, SWA12k-4B-RL-2500 reaches 87.1\%, only 1 points below SA-4B-RL-2200 at 88.1\%. These results suggest that RL can further narrow the gap between SA and SWA at larger scales, consistent with our findings at 1.5B.

\begin{table}[t]
    \centering
    \small
    \caption{Experiments with 4B-parameter models. Math Avg denotes average accuracy across mathematical reasoning benchmarks, and Gap is the difference in Math Avg between SA and SWA. Higher Math Avg is better, and smaller-magnitude gaps indicate that SWA is closer to SA.}\label{tab:4b_results}
    \begin{tabular}{l|ccc}
        \toprule
        Model             & \aofb{T}{eval} & Math Avg & Gap      \\
        \midrule
        SA-4B-SFT         & 19.6           & 84.1     & baseline \\
        SWA12k-4B-SFT     & 11.0           & 80.2     & -3.9     \\
        \midrule
        SA-4B-RL-2200     & 13.4           & 88.1     & baseline \\
        SWA12k-4B-RL-2500 & 6.3            & 87.1     & -1.0     \\
        \bottomrule
    \end{tabular}
\end{table}

\subsubsection{Comparison of SA and SWA on Other Domains}\label{sec:appendix_other_domains}

To further examine how far the main findings extend beyond math reasoning, we compare SA and SWA on a range of additional tasks. Note that these tasks are not the main focus of this paper, and we do not present \emph{SWARR} as a recipe for them.
We divide mature LLM tasks into three main categories according to their input--output lengths: (1) general chat: short input and short output; (2) long-generation reasoning: short input and long generation; and (3) long-context understanding: long input and short output. Tasks with both long input and long output are not yet mature enough to evaluate reliably, so we do not consider them here.

\textbf{Math Reasoning (Short Input and Long Generation)}: We select the best-performing 1.5B checkpoints within 2500 RL steps for this comparison: SA-RL-1300, SWA8k-RL-1500, SWA4k-RL-2200, and SWA2k-RL-1500. As shown in \Cref{tab:other_math}, SWA8k-RL-1500 and SWA4k-RL-2200 achieve math reasoning performance comparable to SA-RL-1300, with average gaps of +0.5 and +0.1 points, respectively.
However, SWA2k-RL-1500 still lags behind SA-RL-1300 by 4.7 points.

\begin{table}
    \centering
    \small
    \caption{
        Comparison of SA and SWA on math reasoning tasks. Math Avg denotes average accuracy across mathematical reasoning benchmarks, and Gap is the difference in Math Avg between SA and SWA. Higher Math Avg is better, and smaller-magnitude gaps indicate that SWA is closer to SA.
    }\label{tab:other_math}
    \begin{tabular}{l|ccccc|c}
        \toprule
        Model         & Math Avg           & Gap      \\
        \midrule
        SA-RL-1300    & \metric{68.1}{0.8} & baseline \\
        \midrule
        SWA8k-RL-1500 & \metric{68.6}{0.8} & +0.5     \\
        SWA4k-RL-2200 & \metric{68.2}{0.7} & +0.1     \\
        SWA2k-RL-1500 & \metric{63.4}{0.8} & -4.7     \\
        \bottomrule
    \end{tabular}
\end{table}

\textbf{General Chat (Short Input and Output)}: We evaluate the above models on general chat ability using language understanding MMLU~\citep{mmlu}, MMLU Pro~\citep{mmlu_pro}, and instruction following IFEval~\citep{ifeval}.
Results are summarized in \Cref{tab:other_general}. All SWA models achieve performance comparable to SA-RL-1300, with average scores within 1.0 point across the three benchmarks. Notably, SWA2k-RL-1500 remains comparable to SA-RL-1300 on general chat tasks even though it lags behind on math reasoning by 4.7 points.

\begin{table}
    \centering
    \small
    \caption{
        Comparison of SA and SWA on general chat tasks. Avg denotes the average score across MMLU, MMLU Pro, and IFEval. Higher scores are better.
    }\label{tab:other_general}
    \resizebox{\linewidth}{!}{
        \begin{tabular}{l|ccc|c}
            \toprule
            Model         & MMLU & MMLU Pro & IFEval & Avg  \\
            \midrule
            SA-RL-1300    & 49.2 & 29.7     & 64.7   & 47.9 \\
            \midrule
            SWA8k-RL-1500 & 49.6 & 29.4     & 63.8   & 47.6 \\
            SWA4k-RL-2200 & 48.9 & 31.6     & 66.4   & 48.9 \\
            SWA2k-RL-1500 & 50.3 & 32.1     & 64.0   & 48.8 \\
            \bottomrule
        \end{tabular}}
\end{table}

\textbf{Long-Context Understanding (Long Input and Short Output)}: Long-context understanding is challenging for SWA because limited memory makes retrieval and evidence integration over long inputs more difficult. Smooth Reading~\citep{smooth_reading} proposes a multi-round agentic inference method that processes long inputs chunk by chunk, helping SWA aggregate information without overwhelming memory. We evaluate SA-RL-1300, SWA4k-RL-2200, and SWA4k-RL-2200-SR (with Smooth Reading) on LongBench~\citep{longbench} and needle-in-the-haystack tasks with 8k and 16k context lengths~\citep{ruler}. Results are summarized in \Cref{tab:long_context_understanding}. Without Smooth Reading, SWA4k-RL-2200 lags behind SA-RL-1300 by 26.24 points on average across the three benchmarks. With Smooth Reading, SWA4k-RL-2200-SR becomes comparable to SA-RL-1300, with an average score within one point.

\begin{table}
    \centering
    \small
    \caption{
        Comparison of SA and SWA on long-context understanding tasks. LB denotes LongBench~\citep{longbench}, and N-8k and N-16k denote needle-in-the-haystack tasks with 8k and 16k context lengths~\citep{ruler}. SWA4k-RL-2200-SR is the SWA4k-RL-2200 model with Smooth Reading~\citep{smooth_reading}. Higher scores are better.
    }\label{tab:long_context_understanding}
    \begin{tabular}{l|ccc|c}
        \toprule
        Model            & LB   & N-8k  & N-16k & Avg   \\
        \midrule
        SA-RL-1300       & 36.0 & 100.0 & 100.0 & 78.7  \\
        \midrule
        SWA4k-RL-2200    & 33.4 & 92.0  & 32.0  & 52.46 \\
        SWA4k-RL-2200-SR & 39.2 & 99.0  & 100.0 & 79.39 \\
        \bottomrule
    \end{tabular}
\end{table}

These results suggest that SWA has potential beyond math reasoning. However, they also show that both the SA--SWA gap and the methods needed to close it vary across domains, suggesting that SWA may require task-specific adaptation to reach its full potential. In this paper, we focus on math reasoning as a representative domain for in-depth analysis and provide an empirical recipe for that setting. We leave broader cross-domain exploration to future work.

\subsubsection{Excluding Ceiling-Effect Explanation for Gap Shrink}\label{sec:appendix_ceiling_effect}

We further rule out a ceiling-effect explanation for the gap shrink with a matched-start control, as shown in \Cref{tab:rl_matched_start}. We introduce SA-Low, an SA model trained with fewer SFT steps so that it starts from a lower accuracy (37.5\%) than SWA8k-SFT (42.5\%). If the gap shrink were caused by SWA starting from a lower accuracy and having more room to improve, the lower-starting SA-Low would be expected to improve more. Instead, after 200 RL steps SA-Low improves by only +10.7 points (to 48.2\%), while SWA8k improves by +15.9 points (to 58.4\%). The gap between SA-Low and SWA8k widens to -10.2\% after RL,
This suggests that the observed gap shrink is not caused by different starting accuracies.

\begin{table}
    \small
    \centering
    \caption{
        Matched-start ceiling control at 200 RL steps. SA-Low is an SA model trained with fewer SFT steps so that it starts from a lower accuracy (37.5\%) than SWA8k-SFT (42.5\%). Despite the lower starting point, SA-Low improves less than SWA8k after RL, suggesting that the gap shrink is not caused by SWA having more room to improve from a lower starting accuracy.
    }\label{tab:rl_matched_start}
    \begin{tabular}{l|c|cc}
        \toprule
        Model         & Setting & Accuracy & Gap      \\
        \midrule
        SWA8k-SFT     & SFT     & 42.5     & Baseline \\
        SA-SFT        & SFT     & 48.6     & +6.1     \\
        SA-Low-SFT    & SFT     & 37.5     & -5.0     \\
        \midrule
        SWA8k-RL-200  & RL      & 58.4     & Baseline \\
        SA-RL-200     & RL      & 59.1     & +0.7     \\
        SA-Low-RL-200 & RL      & 48.2     & -10.2    \\
        \bottomrule
    \end{tabular}
\end{table}

\subsubsection{Gap Shrink under Plain GRPO}\label{sec:appendix_plain_grpo}

We verify that the gap-shrinking trend does not depend on the stabilization techniques by running plain GRPO without any of the additional stabilization components, as shown in \Cref{tab:rl_plain_grpo}. Under plain GRPO with 125 RL steps, SWA8k improves from 42.5\% after SFT to 54.2\%, slightly surpassing SA (53.7\%) and reproducing the gap-shrinking trend of the main experiments. In contrast, the stabilization techniques mainly contribute to training stability and final performance: SA trained with the full stabilization stack reaches 68.1\% (its best checkpoint after convergence, at 1500 RL steps), substantially above the 53.7\% of plain GRPO at 125 steps. We therefore conclude that the stabilization stack allows longer, more stable RL training and stronger absolute performance, while the architecture-conditioned gap-shrinking trend is a more general phenomenon that already appears without it.

\begin{table}
    \small
    \centering
    \caption{
        Plain GRPO without the stabilization stack (125 RL steps). The gap-shrinking trend already appears under plain GRPO, while the stabilization techniques mainly improve training stability and final performance (SA reaches 68.1\% at 1500 RL steps with the full stack).
    }\label{tab:rl_plain_grpo}
    \begin{tabular}{l|c|cc}
        \toprule
        Model        & RL Setting & Accuracy & Gap      \\
        \midrule
        SA-SFT       & SFT        & 48.6     & Baseline \\
        SWA8k-SFT    & SFT        & 42.5     & -6.1     \\
        SA-RL-125    & GRPO       & 53.7     & Baseline \\
        SWA8k-RL-125 & GRPO       & 54.2     & +0.5     \\
        \bottomrule
    \end{tabular}
\end{table}

\subsubsection{On-policy vs.\ Off-policy RL}\label{sec:appendix_on_off_policy}

We further examine the role of architecture-conditioned trajectories by comparing on-policy RL with off-policy RL, as shown in \Cref{tab:rl_on_off_policy}. Off-policy RL is implemented by training SWA on trajectories generated by SA. Under 100 RL steps, on-policy RL reaches 54.0\%, substantially above off-policy RL (49.2\%). On-policy RL is therefore more effective for SWA, supporting the view that architecture-conditioned trajectories are important for SWA's performance.

\begin{table}
    \small
    \centering
    \caption{
        On-policy vs.\ off-policy RL for SWA8k at 100 RL steps. Off-policy RL is implemented by training SWA on trajectories generated by SA. On-policy RL substantially outperforms off-policy RL, supporting the view that architecture-conditioned trajectories are important for SWA's performance.
    }\label{tab:rl_on_off_policy}
    \begin{tabular}{l|c|c}
        \toprule
        Model        & RL Setting & Accuracy \\
        \midrule
        SWA8k-RL-100 & on-policy  & 54.0     \\
        SWA8k-RL-100 & off-policy & 49.2     \\
        \bottomrule
    \end{tabular}
\end{table}

\subsubsection{Reproduction with a Public SFT Dataset}\label{sec:appendix_open_sft}

To examine whether the gap-shrinking trend generalizes across SFT datasets, we repeat the main SFT-to-RL pipeline with the public AceReason math SFT dataset~\citep{liu2025acereason}. As shown in \Cref{tab:ace_open_sft_repro}, after SFT SWA2k trails SA by 4.2 points (36.4 vs.\ 40.6). After 500 RL steps, SWA2k reaches 46.6\%, slightly above SA (46.2\%), reproducing the gap-shrinking trend observed with our internal SFT dataset, although absolute accuracies are lower because the AceReason SFT dataset is smaller and of lower quality than our internal math SFT corpus.

\begin{table}
    \small
    \centering
    \caption{
        Reproduction of the SFT-to-RL pipeline on the public AceReason math SFT dataset~\citep{liu2025acereason}. The same gap-shrinking trend appears as with the internal SFT dataset, although absolute accuracies are lower because this dataset is smaller and of lower quality. Accuracy denotes the average across the mathematical reasoning benchmarks.
    }\label{tab:ace_open_sft_repro}
    \begin{tabular}{l|c|c}
        \toprule
        Model        & Accuracy & Gap (SWA2k $-$ SA) \\
        \midrule
        SA-SFT       & 40.6     & Baseline           \\
        SWA2k-SFT    & 36.4     & $-4.2$             \\
        \midrule
        SA-RL-500    & 46.2     & Baseline           \\
        SWA2k-RL-500 & 46.6     & $+0.4$             \\
        \bottomrule
    \end{tabular}
\end{table}

\subsection{Details of Experiments}\label{sec:appendix_experiment_details}

\subsubsection{Training Datasets}\label{sec:appendix_artifacts}

For training, we use two data sources. The SFT stage uses a math-heavy internal research dataset of 42B tokens, which is not publicly released. Of these, 28B tokens (67\%) are math tokens, drawn from 3.83M question--answer pairs with a mean length of 7{,}211 tokens. Math prompts are sourced from the am-thinking~\citep{am_thinking}, and answers are generated by state-of-the-art LLMs with self-attention. The prompt pool is cleaned with deduplication, strict filtering, and contamination removal against public benchmarks~\citep{am_thinking}. The RL stage uses AceReason-Math~\citep{liu2025acereason}, which contains about 49k mathematical reasoning problems. For evaluation, we use public mathematical reasoning benchmarks in the main paper and additional public benchmarks in the appendix. The corresponding sample counts used in our evaluation are reported in the main paper and appendix.

\subsubsection{Evaluation Details}\label{sec:appendix_eval_details}

During evaluation on math reasoning tasks, we set temperature to 0.6, top-p to 1.0, and top-k to 40 for all models. We extract the answer from the model output by taking the last match of the pattern \texttt{\textbackslash boxed\{answer\}} and compare it with the ground truth using math-verify~\citep{math_verify}.
We report normal-approximation confidence intervals estimated from repeated-evaluation bootstrap samples.
For a benchmark with \(n\) prompts and \(m\) repeated generations per prompt, let \(x_{ij}\in\{0,1\}\) denote whether the \(j\)-th generation for prompt \(i\) is correct.
For each bootstrap trial, we sample one generation \(x_{i,j_i}\) from \(\{x_{ij}\}_{j=1}^{m}\) for every prompt and compute
\begin{equation}
    \small
    \hat{a}^{*} = \frac{1}{n}\sum_{i=1}^{n} x_{i,j_i}.
\end{equation}
We repeat this process for \(B=2000\) trials to obtain bootstrap accuracy estimates \(\{\hat{a}^{*(b)}\}_{b=1}^{B}\).
In tables, \(\metric{a}{h}\) denotes the mean accuracy \(a\) and the half-width \(h=1.96\cdot\mathrm{Std}_{b}(\hat{a}^{*(b)})\) of the normal-approximation 95\% confidence interval.
These intervals summarize repeated-evaluation uncertainty for each model individually. Very small differences whose magnitude is comparable to the reported half-widths should be interpreted as comparable performance rather than strong evidence that one model is better in a given comparison.

\subsubsection{RL Details}\label{sec:appendix_rl_details}

We follow \citet{rl_scaling_law} to implement our RL pipeline.
For clarity, we separate the description into \textbf{core components} (following \citet{rl_scaling_law}) and \textbf{additional stabilization techniques}.
The stabilization techniques are also used in the main experiments, and we further validate their effect in a separate GSM8k ablation with Qwen2.5-0.5B-Instruct.

\paragraph{Core components used in the main experiments.}
Our main RL runs use:
1) the CISPO loss~\citep{Chen2025MiniMaxM1ST};
2) FP32 logits for policy-loss computation;
3) reward normalization by the batch-level reward standard deviation;
4) filtering of groups with zero reward variance;
5) adaptive prompt filtering, where prompts that pass in all repeated evaluations are removed from future sampling; and
6) asynchronous rollout~\citep{fu2025areal}, where optimization can consume rollouts generated by recent previous policy versions to reduce long-tail rollout stalls.

\paragraph{Additional stabilization techniques.}
We further use three stabilization techniques to improve stability and efficiency.

\textbf{Advantage Normalization.}
Large negative advantages can destabilize optimization.
To reduce their magnitude while preserving their sign, we compute a step-level scaling factor
\begin{equation}
    \small
    s=\min\left(
    1,\,
    \frac{\sum_{i=1}^N\sum_{t=1}^T \max(A_{i,t},0)}
    {\epsilon+\sum_{i=1}^N\sum_{t=1}^T \max(-A_{i,t},0)}
    \right),
\end{equation}
where \(N\) is the number of samples in the optimization step, \(T\) is the number of tokens per sample, and \(\epsilon\) is a small constant for numerical stability.
We apply this factor only to negative advantages:
\begin{equation}
    \small
    \tilde{A}_{i,t}=
    \begin{cases}
        A_{i,t},    & A_{i,t}\ge 0, \\
        s\,A_{i,t}, & A_{i,t}<0.
    \end{cases}
\end{equation}

\textbf{Small Entropy Penalty Loss.}
During RL, we observe an increasing number of extremely low-entropy tokens, which can lead to over-exploitation.
For token \(t\), let
\begin{equation}
    \small
    H_{i,t} = -\sum_v p_{i,t,v}\log p_{i,t,v}
\end{equation}
denote the token entropy.
We stop gradients on tokens that are already above the entropy target:
\begin{equation}
    \small
    \tilde{H}_{i,t}=
    \begin{cases}
        H_{i,t},              & H_{i,t}<\tau,   \\
        \mathrm{sg}(H_{i,t}), & H_{i,t}\ge\tau,
    \end{cases}
\end{equation}
where \(\mathrm{sg}(\cdot)\) denotes stop-gradient.
Let
\begin{equation}
    \small
    \bar{H}=\frac{1}{NT}\sum_{i=1}^{N}\sum_{t=1}^{T} \tilde{H}_{i,t}
\end{equation}
be the global mean entropy over the current optimization step.
The entropy penalty is activated only when \(\bar{H}<\tau\):
\begin{equation}
    \small
    \mathcal{L}_{\mathrm{ent}}
    =
    \mathbf{1}[\bar{H}<\tau]\,\lambda
    (\bar{H}-\tau)^2
\end{equation}
Since \(\bar{H}-\tau<0\) when the penalty is active, minimizing this term increases entropy for low-entropy tokens, while tokens already above the target do not receive entropy gradients.
In the main experiments, we set \(\tau=0.3\) and \(\lambda=0.1\).

\textbf{High-Entropy Filtering and Early Stop.}
We also observe occasional high-entropy gibberish outputs.
These outputs may obtain the correct final answer by chance, but their reasoning traces are unreliable.
Therefore, if a rollout exceeds a high-entropy threshold, we mark it as incorrect regardless of answer extraction.
Such outputs are also usually long and can dominate rollout time, so we stop generation early once the entropy of the latest tokens exceeds the threshold.
In the main experiments, the high-entropy threshold is 4.0.

We validate these stabilization techniques by training Qwen2.5-0.5B-Instruct on GSM8k~\citep{gsm8k}, as shown in \Cref{tab:rl_stabilization_ablation}. The baseline consists of the core RL components without the additional stabilization techniques, and also excludes all-pass sample filtering because GSM8k has relatively few training samples. Without these techniques, training diverges at step 400, with a final accuracy of 60.3\%. With advantage normalization, the model remains stable until step 1800 and reaches 63.5\%. With both advantage normalization and entropy penalty, the model stays stable for the full 2500 steps and improves further to 65.1\%. With high-entropy filtering alone, the model also remains stable for 2000 steps and reaches 64.9\%. These results show that the stabilization techniques substantially improve both stability and final performance.

We further verify that the gap-shrinking trend does not depend on these stabilization techniques and that architecture-conditioned trajectories are important for SWA; these analyses are presented in Appendix \Cref{sec:appendix_plain_grpo} and \Cref{sec:appendix_on_off_policy}.

In our main experiments, we tune the RL recipe on the SA model and then apply the same recipe to all SWA models without further tuning. Therefore, the stabilization techniques are not overfit to SWA.
All RL hyperparameters are summarized in \Cref{tab:rl_hparams}.

\begin{table}
    \centering
    \small
    \caption{RL hyperparameters used in the main experiments.}\label{tab:rl_hparams}
    \begin{tabular}{l|c}
        \toprule
        Hyperparameter                         & Value \\
        \midrule
        CISPO high clip (\texttt{h-clip})      & 0.28  \\
        CISPO low clip (\texttt{l-clip})       & 0.20  \\
        Rollouts per prompt                    & 8     \\
        Optimization updates per rollout batch & 4     \\
        Entropy target \(\tau\)                & 0.3   \\
        Entropy penalty weight \(\lambda\)     & 0.1   \\
        High-entropy rejection threshold       & 4.0   \\
        \bottomrule
    \end{tabular}
\end{table}

\begin{table}
    \small
    \centering
    \caption{
        Ablation of RL stabilization techniques on GSM8k with Qwen2.5-0.5B-Instruct. We train for 2500 RL steps and report the final accuracy and the number of steps until the first divergence (defined as a drop of more than 20 points from the previous best accuracy).
    }\label{tab:rl_stabilization_ablation}
    \begin{tabular}{l|cc}
        \toprule
        Method                    & Step  & Gsm8k \\
        \midrule
        Baseline                  & 400   & 60.3  \\
        \midrule
        AdvNorm                   & 1800  & 63.5  \\
        AdvNorm + Entropy Penalty & >2500 & 65.1  \\
        \midrule
        High-Entropy Filtering    & 2000  & 64.9  \\
        \bottomrule
    \end{tabular}
\end{table}

\subsubsection{Infrastructure}\label{sec:appendix_infrastructure}

\textbf{Integrated Inference and Training System.}
Our RL training pipeline, AlloyLM~\citep{alloylm}, is built on XTuner~\citep{2023xtuner}, which is designed for efficient fine-tuning of large language models with Fully Sharded Data Parallel (FSDP)~\citep{zhao2023pytorch}. Instead of using separate systems for inference and training, we implement an integrated system that combines both components.
Inference and training share the same model weights. Specifically, we append generation kernels to the model and implement a simple LLM service within the training process. This design avoids a separate inference engine, making the system lighter and easier to maintain.
The integrated design also makes it easy to implement techniques such as our entropy-based early stop.
For SWA evaluation and timing, we use a true sliding KV cache: keys and values older than the configured window are discarded rather than retained and masked.
Our implementation follows widely used efficient decoding and serving techniques, including FlashAttention~\citep{dao2022flashattention}, continuous batching~\citep{Yu2022OrcaAD}, paged KV-cache management~\citep{kwon2023efficient_vllm}, asynchronous rollout~\citep{fu2025areal}, and the single-controller token-in token-out design of \citet{verl}.

\subsubsection{Responsible Research Statement}
Excluding our own training framework and private SFT dataset, all datasets, benchmarks, pretrained models, and software libraries used in this work are publicly available and are used in accordance with their respective licenses and terms of use.
All datasets we used are used only for research and does not contain personally identifiable information or offensive content.
This paper was edited with the assistance of LLMs for grammar correction, formatting, and language polishing.

\subsection{Impact Statement}

This paper studies a training recipe for efficient long-generation reasoning models rather than a deployed decision-making system. A potential positive impact is reducing the cost of long-generation reasoning research and making experimentation with efficient architectures more accessible.
We view SWARR as a research recipe rather than a deployment recommendation, especially in high-stakes settings where outputs require careful verification.